\documentclass[letterpaper, 10 pt, conference]{ieeeconf}  

\IEEEoverridecommandlockouts                              

\usepackage{graphics} 
\usepackage{epsfig} 
\usepackage{amsmath} 
\usepackage{amssymb}  
\usepackage{psfrag}
\usepackage{textcomp}
\usepackage{tabularx}
\usepackage{stfloats}

\newcolumntype{b}{X}
\newcolumntype{s}{>{\centering\arraybackslash\hsize=.5\hsize}X}

\title{\LARGE \bf
Stiffness-Tuneable Limb Segment with Flexible \\Spine for Malleable Robots
}

\author{Angus B. Clark, \IEEEmembership{Student Member, IEEE} and Nicolas Rojas, \IEEEmembership{Member, IEEE}
\thanks{Angus B. Clark and Nicolas Rojas are with the REDS Lab, Dyson School of Design Engineering, Imperial College London, 25 Exhibition Road, London SW7 2DB, UK. (e-mail: {\tt\footnotesize \{a.clark17, n.rojas\}@imperial.ac.uk}). This work was supported in part by the Engineering and Physical Sciences Research Council grant EP/R020833/1.}%
}

\begin{document}

\maketitle
\thispagestyle{empty}
\pagestyle{empty}

\begin{abstract}

Robotic arms built from stiffness-adjustable, continuously bending segments serially connected with revolute joints have the ability to change their mechanical architecture and workspace, thus allowing high flexibility and adaptation to different tasks with less than six degrees of freedom, a concept that we call malleable robots. Known stiffening mechanisms may be used to implement suitable links for these novel robotic manipulators; however, these solutions usually show a reduced performance when bending due to structural deformation. By including an inner support structure this deformation can be minimised, resulting in an increased stiffening performance. This paper presents a new multi-material spine-inspired flexible structure for providing support in stiffness-controllable layer-jamming-based robotic links of large diameter. The proposed spine mechanism is highly movable with type and range of motions that match those of a robotic link using solely layer jamming, whilst maintaining a hollow and light structure. The mechanics and design of the flexible spine are explored, and a prototype of a link utilising it is developed and compared with limb segments based on granular jamming and layer jamming without support structure. Results of experiments verify the advantages of the proposed design, demonstrating that it maintains a constant central diameter across bending angles and presents an improvement of more than 203\% of resisting force at 180\textdegree.

\end{abstract}

\section{Introduction}

Collaborative robots are a rapidly expanding area of robotics due to their advantages of simplicity in control for a user and safety over classical industrial robots. The majority of these robots typically utilise rigid robotic serial manipulators comprised of links and joints \cite{majidi2014soft}. To allow for complex positioning in reduced spaces and adaptability across given tasks, they commonly integrate six or more degrees of freedom (DOF). However, for the majority of robotic tasks, lower mobility (\emph{i.e.}, less than 6 DOF) is sufficient \cite{yang2000task}. Further, a reduced number of DOF decreases control complexity, weight, and cost of the overall system. Soft robot arms, typically using a continuously bending manipulator, have attempted to provide a solution with reduced size, weight, and increased flexibility. Nevertheless, soft manipulators have been lacking in holding strength and precision in comparison to fully rigid designs \cite{robertson2017soft}. Recent work has explored the use of variable stiffness in soft, continuum robots for surgery, which contain aspects of strength and flexibility by varying the rigidity of the robot \cite{kim2012design, jiang2013stiffness}.

\begin{figure}[!t]
    \centering
    \includegraphics[width=\columnwidth]{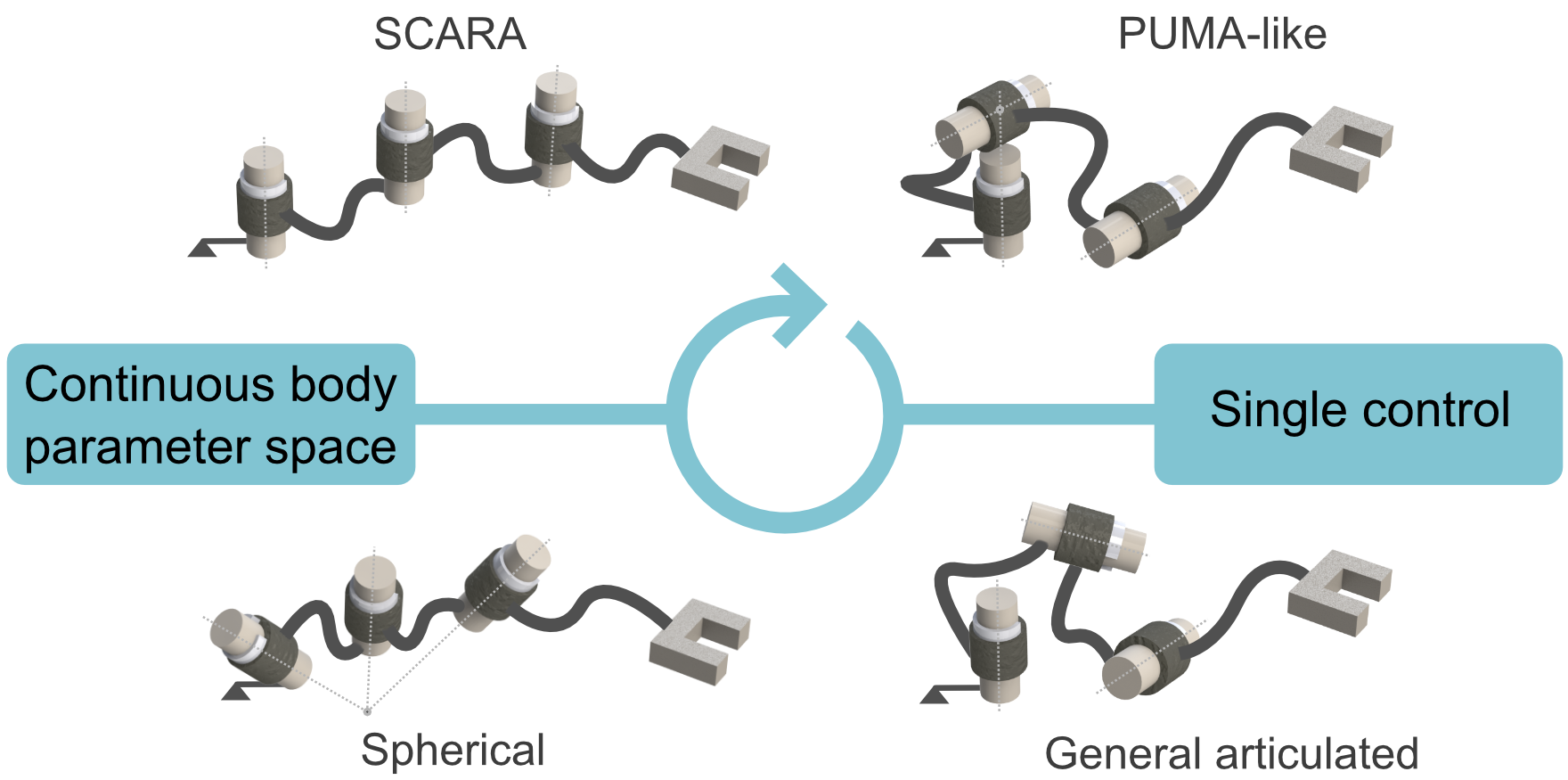}
    \caption{Malleable robots are reconfigurable serial arms of lower mobility (\emph{i.e.}, less than 6 DOF) with variable workspace, able to be adapted to specialised topologies (\emph{e.g.}, SCARA, spherical, or PUMA-like) as well as to other ad-hoc articulated architectures. These robots are composed of stiffness-tuneable, continuously bending limb segments (malleable links) serially connected with revolute joints; they are characterised by having a continuous body parameter space controlled by a single strategy.}  
    \label{MalleableDefinition}
\end{figure}

At present, no serial robots of lower mobility with high flexibility and adaptation to different tasks, that is, with variable but controllable workspace, and able to be adjusted to specialised topologies (e.g., SCARA,  spherical, or PUMA-like) as well as to other ad-hoc articulated  architectures, exist. In the case of confined and remote operation, these robots may facilitate the automation of processes and the augmentation of human operator capabilities, to name some advantages. We define these reduced-DOF reconfigurable serial arms as \emph{malleable robots}, when they are composed of variable stiffness, flexible, continuously bending links which can be reshaped when inactive, while holding its position when activated. This characteristic allows for a variable geometry topology, as shown in Fig. \ref{MalleableDefinition}, where the continuous body parameter space is controlled by a single strategy. Malleable robots can be classified as extrinsic or intrinsic, depending on the source that reshapes their links. In the extrinsic case, we consider this reconfiguration to be performed manually by a human operator to strengthen human-robot collaboration. Malleable robots depart from reconfigurable robotic arm concepts that has been proposed for space applications \cite{Aghil2009}; these systems are composed of lockable, passive joints and traditional rigid links.

Recent research in variable stiffness robotic manipulators has developed a variety of stiffening mechanisms, which have been categorised in \cite{blanc2017flexible}. The majority of variable stiffness technologies have utilised either wire jamming, pneumatic jamming, or structural jamming (such as granular jamming), due to the pre-existing integration of control wires through a manipulator, or the relatively simple use of granules for jamming \cite{ranzani2016soft, jiang2012variable, kim2014stiffness, degani2006highly}. Other methods utilising variable stiffness materials such as magnetorheological and electrorheological fluids, or phase-changing materials such as thermally activated materials, have shown use in robotic applications \cite{pettersson2010design, shintake2015variable, hino2004development, shan2013soft}; however, these approaches show long activation times (in the order of seconds), which makes them inappropriate for the design of limb segments for malleable robots (\emph{i.e.}, malleable links).

A promising structural stiffening method is negative pressure jamming of thin layers of material wrapped along a cylinder, first proposed in \cite{kim2012design}, which shows a maximum lateral resisting force of 2.5N compared to 0.5N for granular jamming when deflected 10mm \cite{langer2018stiffening}. The design also allows for a hollow space through the tubular construction, which is optimal for pass-through of cables, leading to its recent popularity in continuum robotics \cite{santiago2015continuum, langer2018stiffening}. For use as a solution for the development of malleable links, the diameter should be designed for optimal manual manipulation (at 40-50mm \cite{mcdowell2012effects})---much larger than that of continuum robots for minimally invasive surgery (average 5-15mm) \cite{zhou2013robotics}.

A limitation of layer jamming is the stiffness dependence on layer orientation. As the cylindrical design bends, the structure deforms and the resisting force when stiff is decreased. Research has been performed on rigid continuum manipulators, which inherently avoid this problem \cite{yoon2009flexible, mehling2006minimally, blessing2004novel, kim2014stiffness}. However, no research has been reported in the literature on the use of a support structure in conjunction with layer-jamming-based variable stiffness technologies to reduce structural deformation, whilst maintaining the advantages of the latter (\emph{e.g.}, a hollow and light structure). The use of an actuation system with variable stiffness has been proposed, but only as to control the positioning of the continuum manipulator, and not to reduce deformation \cite{langer2018stiffening}. Further, the majority of proposed backbone-like systems are limited in their ability to compress and extend, thus the adaptation of such mechanisms for a support structure would hinder the higher DOF ability of the variable stiffness design.

\begin{figure*}[ht]
    \centering
    \includegraphics[width=0.9\textwidth]{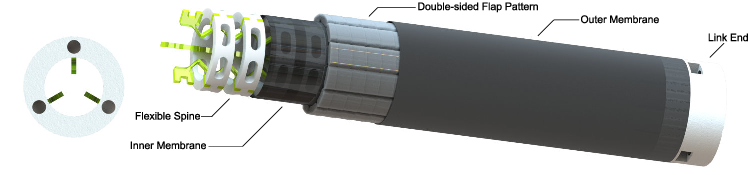}
    \caption{Sectional view of the developed controllable stiffness link with integrated flexible spine for malleable robots {\bf(right)}. Close-up view showing internal layout and hollow centre {\bf(left)}.}
    \label{MalleableLink}
\end{figure*}

In this paper, as a solution to implement malleable links, we propose a unique design for achieving a multi-DOF neutral-line constant diameter support mechanism, namely a flexible spine, to reduce structural deformation in stiffness-controllable layer-jamming-based robotic links of large diameter. The proposed spine mechanism, which is constructed from 3D printed acrylonitrile butadiene styrene (ABS) and thermoplastic polyurethane (TPU) material, is highly movable with type and range of motions that match those of a robotic link using solely layer jamming, without sacrificing its advantages. Moreover, while the resisting force of variable stiffness manipulators at various positions has been previously studied in a number of works (\emph{e.g.}, \cite{langer2018stiffening, jiang2012design}), to the best of the authors' knowledge, this is the first time that the central diameter variation across bending angles of variable stiffness mechanisms and its implications are studied.

The rest of this paper is organised as follows. Section \ref{sec:malleablelinks} covers the design and development of the flexible spine and variable stiffness mechanism to create in combination a malleable link. Section \ref{sec:evaluation} describes the testing methodology and comparison to solutions using granular and layer jamming designs without support structure. To conclude, section \ref{sec:results} and \ref{sec:discussion} presents the results and provides analysis of the introduced malleable link.

\section{Development of Malleable Links} \label{sec:malleablelinks}
\subsection{Layer jamming sheath}
\begin{figure}[t]
    \centering
    \includegraphics[width=0.95\columnwidth]{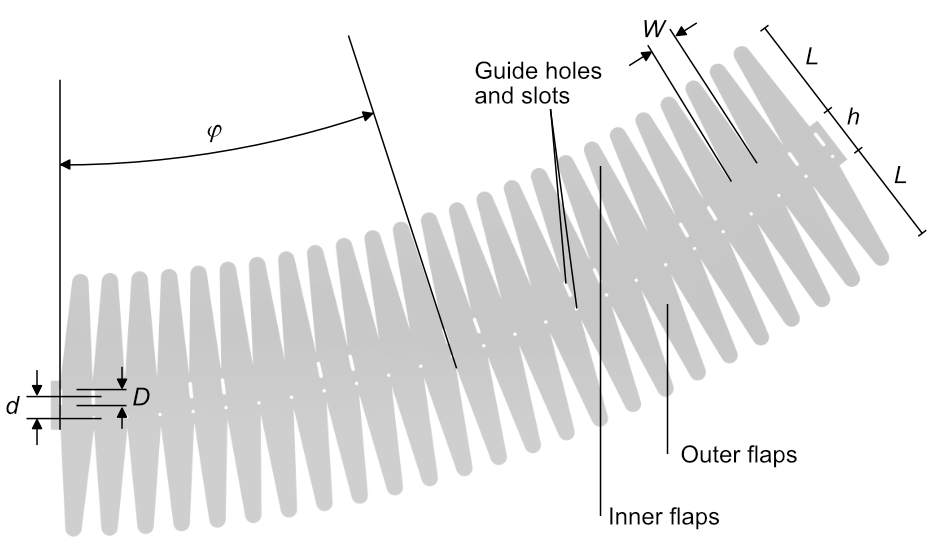}
    \caption{Double-sided flap pattern specifications for layer jamming sheath with guide holes and slots.}
    \label{LayerDesign}
\end{figure}

Our approach to link design for extrinsic malleable robots focused on leveraging existing layer jamming technology used in continuum manipulators to create a finite length variable stiffness link, capable of manual human deformation. The outer stiffening sheath was then based on the double-sided flap pattern presented in \cite{kim2012design}, with the parameters flap length $L$, flap width $W$, mid length $h$, guide hole distance $d$, and inclination angle $\varphi$, detailed in Fig.~\ref{LayerDesign}. The flap pattern is wrapped into an overlapping conic tube, which when compressed between two membranes becomes rigid under the cumulative friction. Compared to the original flap parameters, the overall diameter of the sheath was increased from 20mm to 38mm. We evaluated the layer properties to be suitable for a sheath with a large diameter. The width and length of the flaps determine the number of overlapping layers---7 for each side of the sheath in our design, which determine the overall area providing friction. This determines the maximum stiffness, which can be calculated from the force $F$ required to separate the layers as $F=\mu nPWL$,
where $\mu$ is the coefficient of friction, $n$ is the number of overlapping layers, and $P$ is the applied pressure. By varying the pressure, it is possible to tune the stiffness of the link.

The layers were constructed by laser cutting matte surfaced polyethyleneterephthalate film (Mylar), with a thickness of 0.18mm and a measured coefficient of friction of 0.4. A high stiffness material was selected to ensure the layers did not fail and buckle when a bending force was applied. To construct the sheath, an extra strong polyester thread (G\"utermann M 782) with a diameter of 0.20mm was sewn using a needle along the guide holes, creating a full loop every 12 holes. As the layers wrapped around forming a hollow cylinder, the guide slots of the previous loop were sewn to the following loop every 4 holes, forming connections every 120\textdegree. 

The slot length and guide hole distance $d$ determine the overall maximum and minimum length of the link, and as such the bending ability of the link (without deforming). The maximum bending angle $\theta$ can be calculated as $\theta = (N-1)\sinh{\frac{D}{\phi}}$,
%
 where $N$ is the number of layer loops, $D$ is the slot length, and $\phi$ is the diameter of the layer jamming sheath. The maximum length $l_{max}$ and minimum length $l_{min}$ of the link can be computed as $l_{max},l_{min} = (N-1)(d\pm\frac{D}{2})+h$.
%
%
The link parameter values used can be seen in Table~\ref{layerProperties}, from which it can be computed that $l_{max}$ is 263mm and $l_{min}$ is  139mm, with a default length $l$ of 201mm. The maximum bending angle, $\theta$, is 187\textdegree.

\begin{table}[!t]
\caption{Specifications of Layer Jamming Pattern Parameters}
\label{layerProperties}
\begin{center}
\begin{tabularx}{.9\columnwidth}{bs}
\hline
Specification               &   Value   \\
\hline
Width of flaps ($W$)        &   10mm    \\
Length of flaps ($L$)       &   30mm    \\
Length of mid   ($h$)       &   15mm    \\
Guide hole distance ($d$)   &   6mm   \\
Slot length ($D$)           &   4mm     \\
Sheath diameter ($\phi$)    &   38mm    \\
No. of layer loops ($N$)    &   32      \\
No. of flaps per section    &   12      \\
Contact surfaces per side   &   7       \\
Frictional Coefficient      &   0.40    \\
Vacuum Pressure             &   -60kPa  \\
\hline
\end{tabularx}
\end{center}
\end{table}

To compress the layers together and activate the stiffening mechanism, an outer and inner tubular membrane were sealed at each end around the sheath, forming a sealed tubular volume which could be evacuated compressing the layers together. The membranes were manufactured manually by cutting sheet latex (0.25mm thickness) to the required length and perimeter, then sealing the latex into tubular membranes using liquid latex adhesive. As the design aim was to produce a link, rather than a continuum manipulator, link ends were created to form a finite length for mounting. These link ends were fabricated by 3D printing---fused deposition modeling (FDM) using ABS, and incorporated a mounting position for both the membranes and the layers, which were attached using Ethyl 2-cyanoacrylate (Super Glue). Further, one of the two link ends enabled a connection via a 6mm PVC tubing to a vacuum pump, allowing access to the sealed volume.

\subsection{Flexible spine}

\begin{figure}[!t]
    \centering
    \includegraphics[width=0.95\columnwidth]{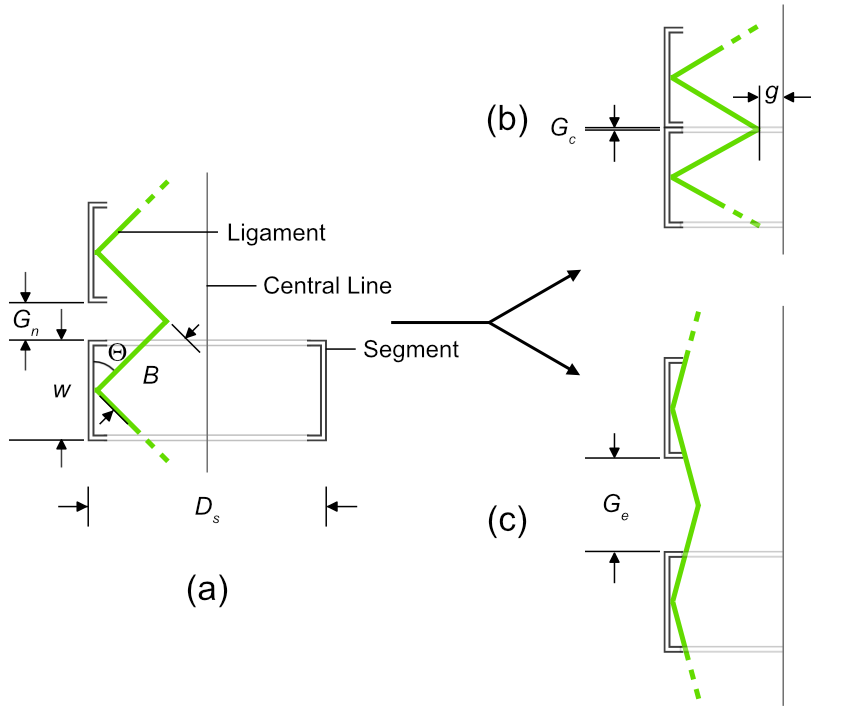}
    \caption{Neutral spine position {\bf (a)}, compressed spine {\bf (b)}, and extended spine {\bf (c)}.}
    \label{SpineDesign}
\end{figure}

To prevent the deformation of the layers when bending a flexible spine was fabricated, which connected to each of the link ends, and maintained the diameter along the link as it underwent bending. The spine properties were identified from the layer jamming sheath, to ensure the spine did not impact the existing bending performance. The spine is defined by the parameters neutral gap ($G_n$), compressed gap ($G_c$), extended gap ($G_e$), ligament beam length ($B$), ligament neutral angle ($\Theta$), central gap ($g$), segment height ($w$), and segment diameter ($D_s$), as shown in Fig.~\ref{SpineDesign}. To ensure the spine DOF did not affect the layers an equal or lower rigid:flexible ratio of component lengths was required, to allow the spine equal or greater compression and extension than the sheath. From the maximum and minimum link lengths ($l_{max}$, $l_{min}$) a ratio of rigid:flexible of 139:62 was obtained. This was computed assuming the flexible components were capable of compressing to a length of zero, and extending to the required length of $l_{max}-l_{min}=124\textrm{mm}$, implying an increase of 100\% from rest position. The spine was constructed from 14 rigid ABS segments, each 8mm in height and 32.5mm in diameter, totalling to 132mm of rigid support when added to the 10mm of support provided by each link end.

\begin{figure}[!t]
    \centering
    \includegraphics[width=\columnwidth]{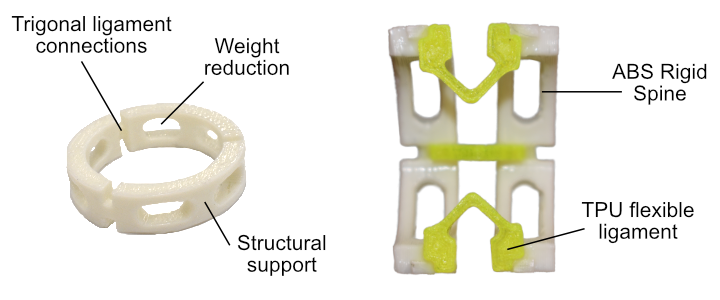}
    \caption{ABS rigid spine segment {\bf (left)} and sectional view of 2-segment spine with integrated flexible ligaments (square planar ligament connections) {\bf (right)}.}
    \label{SpineCross}
\end{figure}

Connecting the segments were 3 magazine spring-inspired TPU ligaments, which were fixed every 120\textdegree\ to the segments and link ends using Ethyl 2-cyanoacrylate. The spine was fabricated via 3D printing, with rigid components formed from ABS and flexible components from TPU. The rigid segment design, as well as a cross-section demonstrating the construction of the segments with the ligaments, is shown in Fig.~\ref{SpineCross}. The flexible ligaments were composed of connectors and triangular springs. The triangular springs connected to the centre of the rigid segment, leaving a neutral gap of 4.53mm between each segment. The spring could compress to reduce the gap to zero, or expand to a maximum of 11mm. Combined with the rigid segments the spine had an overall length of 202mm, and could compress to 132mm and expand to 297mm, ensuring it did not impact the sheath performance. When fully compressed, the central gap $g$ can be computed as
\begin{equation}
    \label{central-gap}
    g = \frac{D_s}{2}-(B^2-(G_c+\frac{w}{2})^2).
\end{equation}

\begin{figure}[t!]
    \centering
    \includegraphics[width=0.8\columnwidth]{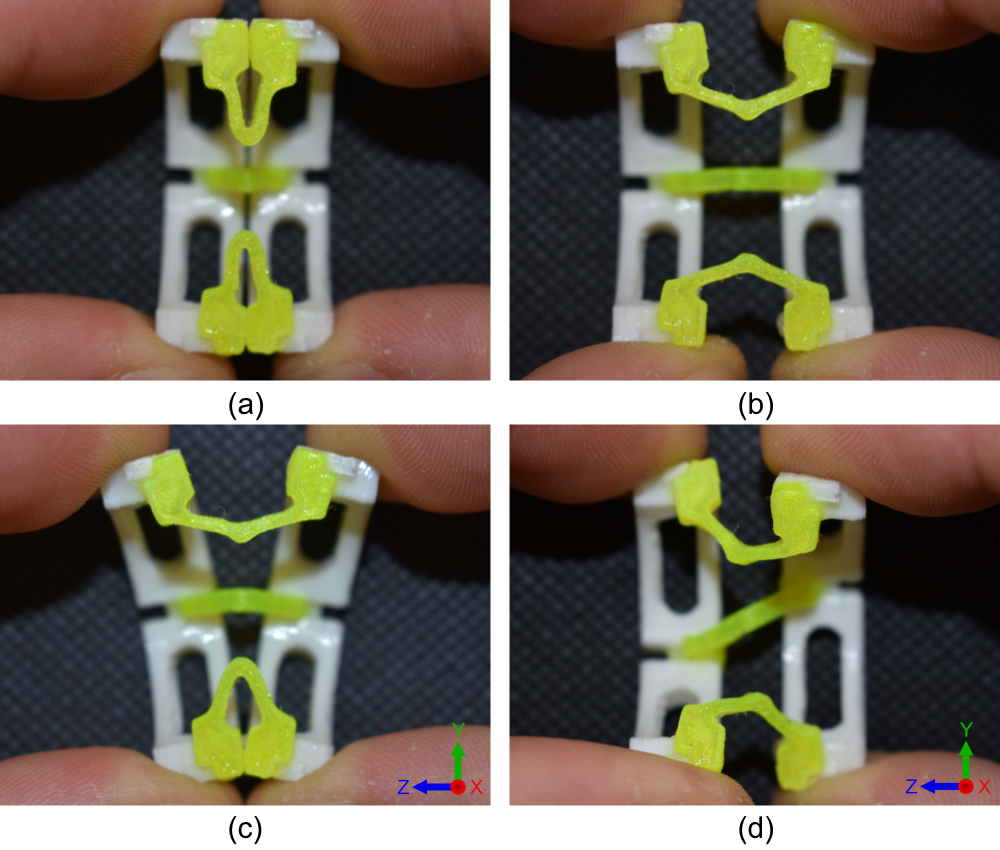}
    \caption{Motion of the flexible spine. Compression {\bf (a)}, extension {\bf (b)}, bending {\bf (c)}, and twist {\bf (d)}.}
    \label{SpineDOF}
\end{figure}

With the compression and expansion limits known, we can rearrange equation \eqref{central-gap} to solve for ligament beam length, substituting the central gap for $\geq$7.5mm, to allow for a full diameter of $\geq$15mm to ensure adequate cable pass-through. Solving, this gives a ligament beam length of $\leq$9.6mm. The selected value for the ligament beam length was 9mm, and the central gap was calculated as 8.2mm. The ligament neutral angle was set as 45\textdegree\ to allow for equal support to be generated at the stress concentrations of the ligament. The motion ability of the spine is demonstrated in Fig.~\ref{SpineDOF}.

\section{Evaluation of Malleable Link Performance}\label{sec:evaluation}

\begin{figure}[t!]
    \centering
    \includegraphics[width=\columnwidth]{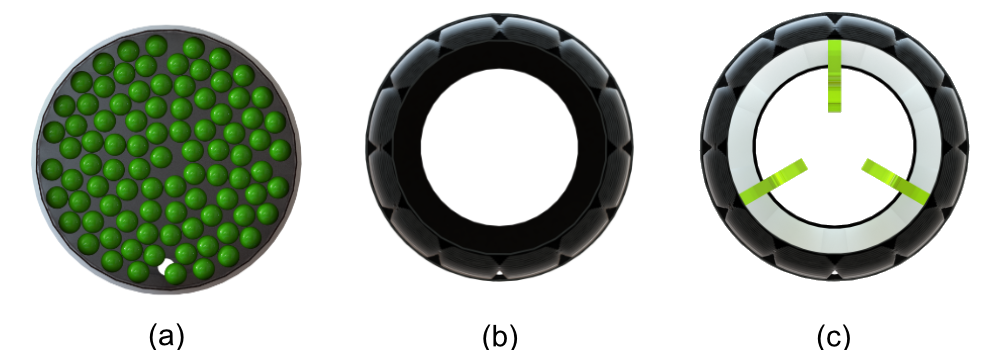}
    \caption{Cross-sectional view of malleable link designs. Granular jamming without support structure {\bf (a)}, layer jamming without support structure {\bf (b)}, and layer jamming with flexible spine {\bf (c)}.}
    \label{LinkCross}
\end{figure}

As no previous work had been performed on similar diameter variable stiffness solutions, to evaluate the performance of our malleable link design a further two malleable links without support structure, based on granular jamming and layer jamming, were constructed and compared. The cross sections of the three malleable links can be seen in Fig. \ref{LinkCross}. The granular jamming link was filled using 4mm diameter matt surfaced glass beads, and sealed using the same outer membrane described in section \ref{sec:malleablelinks}. The layer jamming design was equivalent to our malleable link design, except without the inclusion of a spine.

In order to evaluate the performance of the malleable links, first their resisting force in various positions were measured. The links were mounted horizontally at one end, and a linear actuator (Actuonix L12-100-100-12I) pushed a load cell (DBBSM 5kg S-Beam) into the end of the link, at an angle perpendicular to the mounted link end, deflected the link 10mm before returning the sensor to its original position. The applied vacuum pressure to each of the links when active was -60kPa (41.9kPa absolute pressure), achieved using an oil-based vacuum pump (BACOENG 220V/50Hz BA-1 Standard), and was measured and controlled using a vacuum regulator (SMC IRV10-C06). The resulting data was measured electronically using LabView and a datalogger.

\begin{figure*}[!t]
    \centering
    \includegraphics[width=0.9\textwidth]{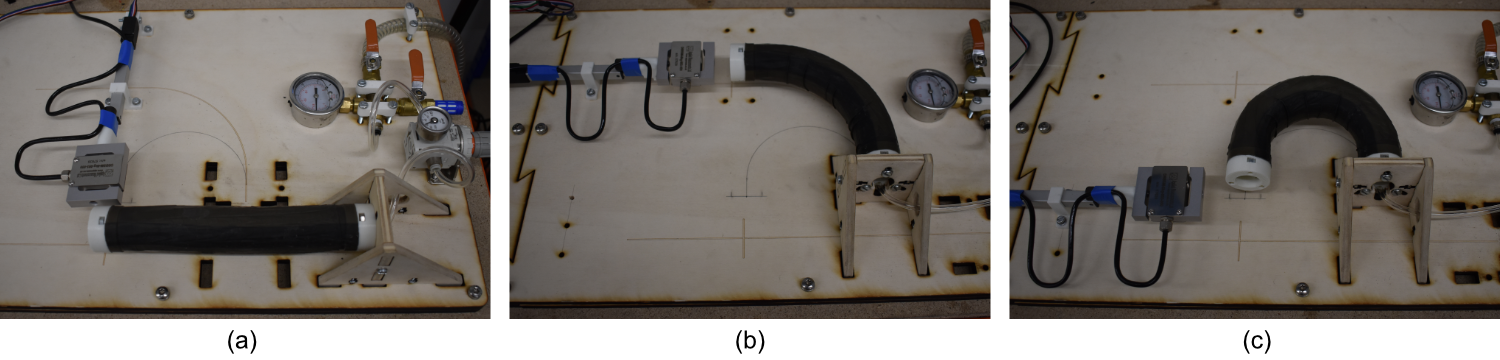}
    \caption{Experiment setups to evaluate the resisting force of the links with 10mm deflection at 0\textdegree\ bending {\bf (a)}, 90\textdegree\ bending {\bf (b)}, and 180\textdegree\ bending {\bf (c)}.}
    \label{ResistingForce}
\end{figure*}

Three bending positions were tested to evaluate the resisting force, namely at 0\textdegree, 90\textdegree, and 180\textdegree as shown in Fig.~\ref{ResistingForce}. The testing of a link at each position was repeated 10 times to ensure accurate recordings. Before a link was stiffened for testing, its position was checked using a guide laser etched onto the testbed. This ensured each link in any position had a length of 220mm, and was horizontal before testing. Once stiffened, the guide was removed, and the test was carried out. For the 90\textdegree\ and 180\textdegree\ experiments the link was bent into position manually.

Moreover, to test further the performance of the malleable links, their central diameter was measured as each link was bent manually from 0\textdegree\ to 180\textdegree\ in 45\textdegree\ increments. At each increment the central diameter was photographed and measured digitally. 

\section{Experiment Results}\label{sec:results}
\subsection{Force displacement}
\begin{figure}[t!]
    \centering
    \includegraphics[width=\columnwidth]{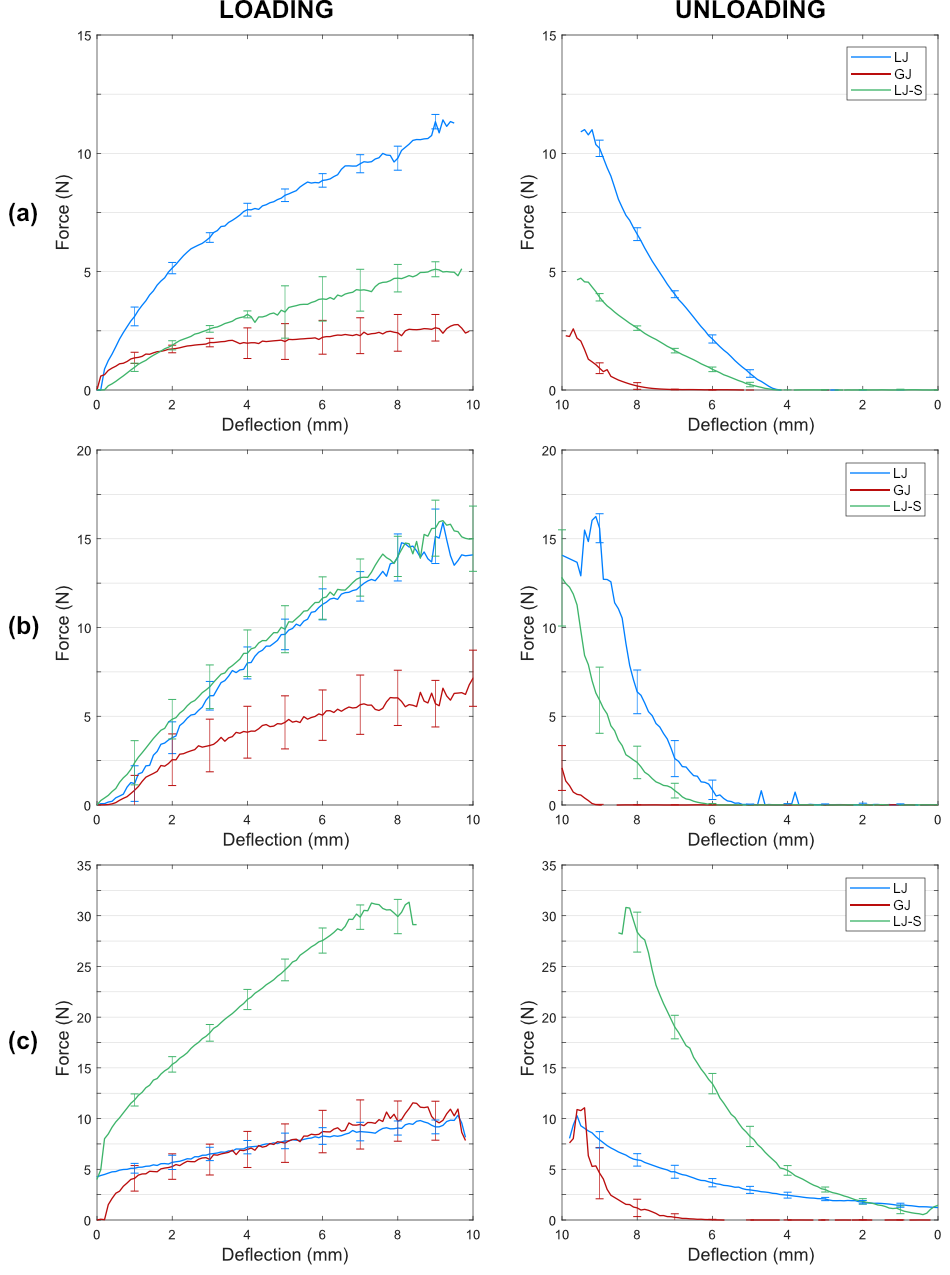}
    \caption{Force-displacement plots for 0\textdegree\ {\bf (a)}, 90\textdegree\ {\bf (b)}, and 180\textdegree\ {\bf (c)} bending tested at absolute pressure 41.9kPa. Granular jamming without support structure (GJ) [\textit{red}], layer jamming without support structure (LJ) [\textit{blue}], and layer jamming with flexible spine (LJ-S) [\textit{green}].}
    \label{ResultsResistingForce}
\end{figure}

The results show the measured force for the given deflections of each link, and were calculated from the mean of the ten repeated experiments per link. Error bars showing $\pm$1 standard deviation every 1mm are included on the force-displacement graphs. The recorded force-displacement plots for the granular jamming, layer jamming, and layer jamming with flexible spine experiments at 0\textdegree, 90\textdegree, and 180\textdegree\ are shown in Fig.~\ref{ResultsResistingForce}. The plots are aligned such that when the load cell first makes contact with the link, deflection is $\approx$ 0mm---after which the force increases due to the resisting force exerted by the link as it is deflected. When the load cell is retracted, the force decreases until either the link maintains a deflection and the force goes to 0N (hysteresis), or until the load cell is at its original 0mm position. A lower hysteresis is preferred to prevent a positional change of the link.  For the 180\textdegree\ experiment, the layer jamming based links were unable to hold their position after activation due to the low vacuum pressure, and as such exerted a force on the load cell at 0mm deflection.

Initially, at 0\textdegree\ bending, the layer jamming  and layer jamming with flexible spine links resulted in a maximum force of 11.41N and 5.12N, respectively. At 90\textdegree\ bending, the layer jamming and layer jamming with flexible spine links demonstrated similar performance, with maximum resisting forces of 15.92N and 16.02N, respectively. When undergoing a full 180\textdegree\ bend, the layer jamming link resisting force significantly decreased to 10.33N due to buckling of the link, compared to 31.33N of resisting force exerted by the layer jamming link with flexible spine. The granular jamming link consistently showed the lowest resisting force without buckling, at 2.76N, 7.14N, and 11.55N, for 0\textdegree, 90\textdegree, and 180\textdegree, respectively.

\subsection{Central diameter variation}
\begin{figure}[t]
    \centering
    \includegraphics[width=\columnwidth]{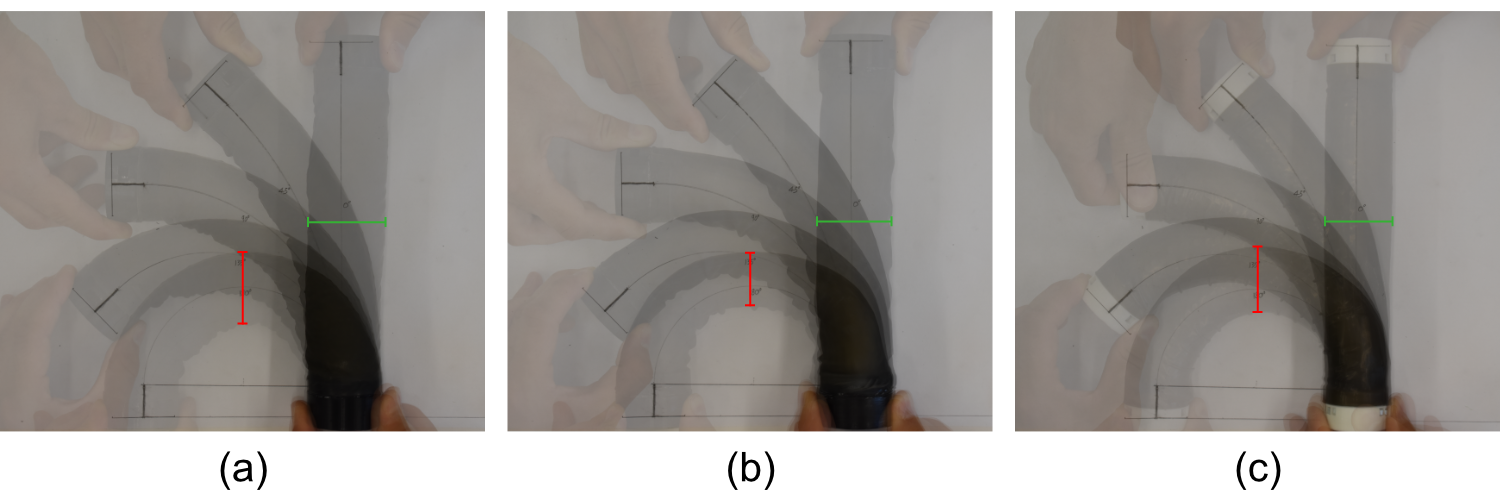}
    \caption{Central diameter variation with bending angle. Granular jamming {\bf(a)}, layer jamming {\bf(b)}, and layer jamming with flexible spine {\bf(c)}.}
    \label{ResultsDiameter}
\end{figure}

The effect of the bending angle on the central diameter of the link can be seen in Fig.~\ref{ResultsDiameter} and Fig.~\ref{ResultsCentralDiameter}. The images composing Fig. \ref{ResultsDiameter} were measured digitally to obtain the central diameter, which was compared against the diameter when at 0\textdegree\ (highlighted green on Fig.~\ref{ResultsDiameter}). The maximum bending angle of 180\textdegree\ is also highlighted to indicate the maximum central diameter variation (in red).


The resulting central diameter ratios were then plotted against bending angle, shown in Fig.~\ref{ResultsCentralDiameter}. The layer jamming link displayed the highest change in diameter, as it exhibited buckling at high bending angles, resulting in a 0.745 ratio at 180\textdegree . Granular jamming initially presented a decrease in diameter up to 90\textdegree; however, despite not containing an internal structure, maintained its diameter from 90\textdegree\ to 180\textdegree, showing a minimum ratio of 0.929. The layer jamming with spine link demonstrated the highest central diameter ratio, resulting in a ratio of 0.979 at 180\textdegree.


\section{Discussion}\label{sec:discussion}
\begin{figure}[t]
    \centering
    \includegraphics[width=0.9\columnwidth]{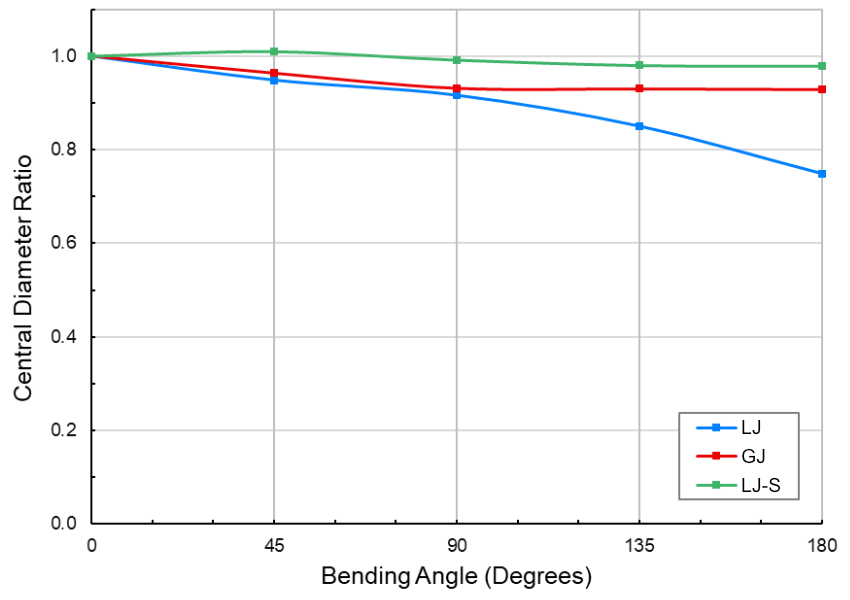}
    \caption{Central diameter ratio-bending angle curves of malleable links: granular jamming (GJ), layer jamming (LJ), and layer jamming with flexible spine (LJ-S)}
    \label{ResultsCentralDiameter}
\end{figure}

From the results shown in Fig. \ref{ResultsResistingForce}, the resisting force experiments exhibit an increase in resisting force with an increase in bending angle for all of the malleable links except the layer jamming link without support structure at 180\textdegree . This can be explained by link buckling at its central diameter point, causing the link to fail and the resisting force to significantly decrease. This can be observed in the central diameter variation in Fig.~\ref{ResultsCentralDiameter}, where the central diameter of the layer jamming can be seen to begin to buckle as the ratio reduces as the bending angle increases. The layer jamming link with flexible spine rectified this issue, ensuring the sheath did not buckle at high bending angles, indicated by the significant reduction in central diameter variation over bending angle. However, surprisingly, our design performed poorly at 0\textdegree, showing only a maximum force of 5.12N, compared to 11.41N from sole layer jamming, a loss of 45\%.

The above result can be explained by the manufacturing process used in the layer jamming designs to ensure that the layers did not damage the membranes at the link ends. The flap length towards each end of the sheaths was reduced, ending with flap lengths of 0mm. However, when attached to the link ends, multiple layers may attach due to adhesive seeping between the layers. In the case of the link with flexible spine, less layers were attached, thus affecting the performance exhibited at 0\textdegree\ bending. This is indeed confirmed by the equal resisting force at 90\textdegree\ bending for both layer jamming based designs, where the force can not concentrate on the weak end of the link. We would expect the design with solely layer jamming to perform with a similar force of $\sim$30N at 180\textdegree\ as the design with flexible spine if it did not buckle under the bending, and an even better result for the flexible spine design without a weaker link end. The granular jamming design performed consistently, however did not display a resisting force as high as any of the non-buckled layer jamming designs. All designs present hysteresis after deformation, with granular jamming performing worse, and both layer jamming designs at 180\textdegree\ acting elastically due to buckling and bending limitations.

The central diameter is an important measure for malleable robotics as one of the aims of extrinsic malleable robots is the ability to be manually reshaped. If the central diameter is capable of changing when bent, and is susceptible to buckling under forces acting at extreme or non-standard angles (such as mid-link), then the link is limited in maximum exertable force by a user. As a user grasps the link and manually shapes the link, they exert a non-normal force on the link, which without support can cause the link to structurally collapse. With a constant diameter, we can ensure the effect of non-normal forces on the link do not lead to a structural collapse, and that the link will perform as the user expects when undergoing manual shaping. Further, the closer to a cylindrical diameter the sheath exhibits, the more reliable its performance is in relation to forces at angles around the link.

\section{Conclusion}
We introduced herein a new category of robots, called malleable robots, defined as reduced DOF reconfigurable serial arms, composed of variable stiffness, continuously bending links and revolute joints. We presented a malleable link capable of maintaining structural integrity while undergoing bending. The malleable link was based on layer jamming, with the inclusion of a flexible 3D printed flexible spine for support. The spine ensured the link was not prone to buckling, and to the best of our knowledge the layer jamming sheath is the largest in diameter presented so far at 38mm. We evaluated the performance of our design, compared to malleable link prototypes based on solely layer jamming and granular jamming. Our presented design performed best, showing a maximum resisting force of 31.33N when bent to 180\textdegree, an increase of 203.3\% over layer jamming and 171.3\% over granular jamming. The prototype was also evaluated on central diameter variation, achieving the lowest variation in central diameter with a ratio of 0.979, a percentage change of only 2.1\% compared to 25.5\% for layer jamming at 180\textdegree. Our design also showed increased malleability over other designs due to its lack of buckling when manually repositioned, allowing more force to be exerted by a user when handling and positioning.

Overall, the results show that the proposed malleable link design is a promising approach toward the creation of malleable collaborative robots. Further improvements to the design will be explored, such as identifying the optimal diameter for both manual manipulation and providing variable stiffness. Moreover, the combination of variable stiffness technologies together to provide advantages of each may also be explored, in order to improve stiffening ability in relation to stiffness, hysteresis, linearity, and stiffness range across bending angles.

\addtolength{\textheight}{-11cm}   






\end{document}